\documentclass[british,runningheads]{llncs}
\usepackage[T1]{fontenc}
\usepackage[latin9]{inputenc}
\usepackage{array}
\usepackage{wrapfig}
\usepackage{booktabs}
\usepackage{multirow}
\usepackage{amsmath}
\usepackage{graphicx}

\makeatletter

\providecommand{\tabularnewline}{\\}
\newcommand{\lyxdot}{.}

\usepackage{makeidx}
\usepackage{graphicx}
\usepackage{amsmath,amssymb} 
\usepackage{color}
\usepackage{listings}

\usepackage{xspace}

\@ifundefined{showcaptionsetup}{}{%
 \PassOptionsToPackage{caption=false}{subfig}}
\usepackage{subfig}
\makeatother

\usepackage{babel}
\begin{document}
\pagestyle{headings} 
\mainmatter

\newcommand{\dpm}{DPM-Hinge\xspace}
\newcommand{\dpmvp}{DPM-Hinge-VP\xspace}
\newcommand{\dpmvoc}{DPM-VOC\xspace}
\newcommand{\dpmvocvp}{DPM-VOC+VP\xspace}
\newcommand{\dpmvocandvp}{DPM-VOC(+VP)\xspace}
\newcommand{\dpmvpcont}{DPM-VP-Cont\xspace}
\newcommand{\dpmthreedparts}{DPM-3D-Constraints\xspace}
\newcommand{\dpmthreed}{3D$^2$PM\xspace}
\newcommand{\dpmthreedd}{3D$^2$PM-D\xspace}     
\newcommand{\dpmthreedc}{3D$^2$PM-C\xspace}
\newcommand{\dpmthreedclin}{3D$^2$PM-C-Lin\xspace}
\newcommand{\dpmthreedcexp}{3D$^2$PM-C-Exp\xspace}
\newcommand{\vdpm}{VDPM\xspace}
\newcommand{\rcnnmv}{RCNN-MV\xspace}
\newcommand{\cnnmv}{CNN-MV\xspace}
\newcommand{\cnn}{convnet}
\newcommand{\Cnn}{Convnet}
\newcommand{\rcnnv}{RCNN-V\xspace}
\newcommand{\rcnnRidge}{RCNN-Ridge\xspace}
\newcommand{\rcnnLasso}{RCNN-Lasso\xspace}
\newcommand{\rcnnElNet}{RCNN-ElNet\xspace}
\newcommand{\rcnnl}{RCNN-L\xspace}
\newcommand{\rcnnvl}{RCNN-V-L\xspace}
\newcommand{\rcnnRidgeL}{RCNN-Ridge-L\xspace}
\newcommand{\rcnn}{RCNN\xspace}
\newcommand{\keyReg}{RCNN-KeyReg\xspace}

\newcommand{\aavp}{AAVP\xspace}
\newcommand*\NewPage{\newpage\null\thispagestyle{empty}\newpage}
\newcommand{\argmax}{\operatornamewithlimits{argmax}}
\newcommand{\argmin}{\operatornamewithlimits{argmin}}
\newcommand{\mcalN}{\mathcal{N}}
\newcommand{\mcalY}{\mathcal{Y}}
\newcommand{\sbt}{\mbox{sb.t.}}
\newcommand{\VOCloss}{\Delta_{\tiny\mbox{VOC}}}
\newcommand{\VOCVPloss}{\Delta_{\tiny\mbox{VOC+VP}}}
\newcommand{\VPloss}{\Delta_{\tiny\mbox{VP}}}
\newcommand{\tu}{\text{u}}
\newcommand{\tv}{\text{v}}
\newcommand{\tx}{\text{x}}
\newcommand{\ty}{\text{y}}
\newcommand{\tz}{\text{z}}
\newcommand{\pgcomment}[1]{\textcolor{red}{pg: #1}}
\newcommand{\pgsuggest}[1]{\textcolor{red}{pg-suggest: #1}}
\newcommand{\mstcomment}[1]{\textcolor{green}{mst: #1}}
\newcommand{\bpcomment}[1]{\textcolor{blue}{bp: #1}}
\newcommand{\bscomment}[1]{\textcolor{blue}{BS: #1}}

\newcommand{\bq}{\mathbf{q}}
\newcommand{\bp}{\mathbf{p}}
\newcommand{\R}{\mathbb{R}}

\newcommand{\todo}[1]{\color{red}{TODO: #1}}
\newcommand{\myparagraph}[1]{{{\vspace{0.1cm}\textbf{#1}\quad}}}
\newcommand{\scream}[1]{{\bf *** #1! ***}}

\title{What is Holding Back \Cnn s for Detection?\vspace{-0.5em}
}

\titlerunning{What is Holding Back \Cnn s for Detection?}

\authorrunning{Bojan Pepik, Rodrigo Benenson, Tobias Ritschel, Bernt Schiele.}

\author{Bojan Pepik, Rodrigo Benenson, Tobias Ritschel, Bernt Schiele}

\institute{Max-Planck Institute for Informatics\vspace{-1.5em}
}
\maketitle
\begin{abstract}
Convolutional neural networks have recently shown excellent results
in general object detection and many other tasks. Albeit very effective,
they involve many user-defined design choices. In this paper we want
to better understand these choices by inspecting two key aspects ``what
did the network learn?'', and ``what can the network learn?''.
We exploit new annotations (Pascal3D+), to enable a new empirical
analysis of the R-CNN detector. Despite common belief, our results
indicate that existing state-of-the-art \cnn~architectures are not
invariant to various appearance factors. In fact, all considered networks
have similar weak points which cannot be mitigated by simply increasing
the training data (architectural changes are needed). We show that
overall performance can improve when using image renderings for data
augmentation. We report the best known results on the Pascal3D+ detection
and view-point estimation tasks. 
\end{abstract}

\makeatletter 
\renewcommand{\paragraph}{%
\@startsection{paragraph}{4}%
{\z@}{1.0ex \@plus 1ex \@minus .2ex}{-0.5em}%
{\normalfont \normalsize \bfseries}%
}
\makeatother

\section{\label{sec:Introduction}Introduction}

\vspace{-0.1em}

In the last years convolutional neural networks (convnets) have become
``the hammer that pounds many nails'' of computer vision. Classical
problems such as general image classification \cite{Krizhevsky2012Nips},
object detection \cite{Girshick2014ArxivRCNN}, pose estimation \cite{Chen2014Nips},
face recognition \cite{Schroff2015ArxifFacenet}, object tracking
\cite{Li2014AccvCnnForTracking}, keypoint matching \cite{Fischer2014ArxivMatching},
stereo matching \cite{Zbontar2015Cvpr}, optical flow \cite{Fischer2015ArxivFlownet},
boundary estimation \cite{Xie2015ArxivEdgeDetection}, and semantic
labelling \cite{Long2015Cvpr}, have now all top performing results
based on a direct usage of convnets. The price to pay for such versatility
and good results is a limited understanding of why convnets work so
well, and how to build \& train them to reach better results.

In this paper we focus on convnets for object detection. For many
object categories \cnn s have almost doubled over previous detection
quality. Yet, it is unclear what exactly enables such good performance,
and critically, how to further improve it. The usual word of wisdom
for better detection with \cnn s is ``larger networks and more data''.
But: how should the network grow; which kind of additional data will
be most helpful; what follows after fine-tuning an ImageNet pre-trained
model on the classes of interest? We aim at addressing such questions
in the context of the R-CNN detection pipeline \cite{Girshick2014ArxivRCNN}
(\S \ref{sec:The-RCNN-detector}).

Previous work aiming to analyse \cnn s have either focused on theoretical
aspects \cite{Bengio2011Alt}, visualising some specific patterns
emerging inside the network \cite{Le2012Icml,Simonyan2014Iclrw,Springenberg2015Iclr,Mahendran2015Cvpr},
or doing ablation studies of working systems \cite{Girshick2014ArxivRCNN,Chatfield2014Bmvc,Agrawal2014Eccv}.
However, it remains unclear what is withholding the detection capabilities
of convnets.

\paragraph{Contributions}

This paper contributes a novel empirical exploration of R-CNNs for
detection. We use the recently available Pascal3D+\cite{xiang14wacv}
dataset, as well as rendered images to analyze R-CNNs capabilities
at a more  detailed level than previous work. In a new set of experiments
we explore which appearance factors are well captured by a trained
R-CNN, and which ones are not. We consider factors such as rotation
(azimuth, elevation), size, category, and instance shape.  We want
to know which aspects can be improved by simply increasing the training
data, and which ones require changing the network. We want to answer
both ``what did the network learn?'' (\S\ref{sec:What-did-the-convnet-learn})
and ``what can the network learn?'' (\S\ref{sec:What-could-the-convnet-learn}
and \S\ref{sec:Which-synthetic-data-helps}). Our results indicate
that current \cnn s (AlexNet \cite{Krizhevsky2012Nips}, GoogleNet
\cite{Szegedy2014ArxivGoogleNet}, VGG16 \cite{simonyan15iclr}) struggle
to model small objects, truncation, and occlusion and are not invariant
to these factors. Simply increasing the training data does solve these
issues. On the other hand, properly designed synthetic training data
can help pushing forward the overall detection performance.

\subsection{\label{sub:Related-work}Related work}

\paragraph{Understanding \cnn s}

The tremendous success of \cnn s coupled with their black-box nature
has drawn much attention towards understanding them better. Previous
analyses have either focused on highlighting the versatility of its
features \cite{Razavian2014Cvprw,Razavian2014ArxivLocalImageProperties},
learning equivariant mappings \cite{lenc15cvpr}, training issues
\cite{Dauphin2014Nips,Ioffe2015Arxiv}, theoretical arguments for
its expressive power \cite{Bengio2011Alt}, discussing the brittleness
of the decision boundary \cite{Szegedy2014Iclr,Goodfellow2015Iclr},
visualising specific patterns emerging inside the network \cite{Le2012Icml,Simonyan2014Iclrw,Springenberg2015Iclr,Mahendran2015Cvpr},
or doing ablation studies of working systems \cite{Girshick2014ArxivRCNN,Chatfield2014Bmvc,Agrawal2014Eccv}.
\\
We leverage the recent Pascal3D+ annotations \cite{xiang14wacv} to
do a new analysis complementary to previous ones. Rather than aiming
to explain how does the network work, we aim at identifying in which
cases the network does not work well, and if training data is sufficient
to improve these issues. While previous work has shown that \cnn\xspace
representations are increasingly invariant with depth, here we show
that current architectures are still not overall invariant to many
appearance factors.

\paragraph{Synthetic data }

 The idea of using rendered images to train detectors has been visited
multiple times. Some of the strategies considered include video game
renderings \cite{Xu2014} (aiming at photo-realism), CAD model wire-frame
renderings \cite{Stark2010Bmvc,bojan15pami} (focusing on object boundaries),
texture-mapped CAD models \cite{kostas14cvpr,Peng14iclr}, or augmenting
the training set by subtle deformations of the positive samples \cite{Enzweiler2008Cvpr,Pishchulin2012Cvpr}.\\
Most of these works focused on DPM-like detectors, which can only
make limited use of large training sets \cite{Zhu2012Bmvc}. In this
paper we investigate how different types of renderings (wire-frame,
materials, and textures) impact the performance of a \cnn. A priori
\cnn s are more suitable to ingest larger volumes of data.

\section{\label{sec:The-RCNN-detector}The R-CNN detector}

The remarkable \cnn\xspace results in the ImageNet 2012 classification
competition \cite{Krizhevsky2012Nips} ignited a new wave of neural
networks for computer vision. R-CNN \cite{Girshick2014ArxivRCNN}
adapts such \cnn s for the task of object detection, and has become
the de-facto architecture for state-of-the-art object detection (with
top results on Pascal VOC \cite{Everingham2007} and ImageNet \cite{deng09cvpr})
and is thus the focus of attention in this paper. The R-CNN detector
is a three stage pipeline: object proposal generation \cite{uijlings13ijcv},
\cnn\xspace feature extraction, and one-vs-all SVM classification.
We refer to the original paper for details of the training procedure
\cite{Girshick2014ArxivRCNN}. Different networks can be used for
feature extraction (AlexNet \cite{Krizhevsky2012Nips}, VGG \cite{Chatfield2014Bmvc},
GoogleNet \cite{Szegedy2014ArxivGoogleNet}), all pre-trained on ImageNet
and fine-tuned for detection. The larger the network, the better the
performance. The SVM gains a couple of final mAP points compared to
logistic regression used during fine-tuning (and larger networks benefit
less from it \cite{Girshick2015ArxivFastRcnn}).

In this work we primarily focus on the core ingredient: \cnn\xspace
fine-tuning for object detection. We consider fine-tuning with various
training distributions, and analyse the performance under various
appearance factors. Unless otherwise specified reported numbers include
the SVM classification stage, but not the bounding box regression.

\section{\label{sec:Pascal-3d-dataset}Pascal3D+ dataset}

Our experiments are enabled by the recently introduced Pascal3D+ \cite{xiang14wacv}
dataset. It enriches PASCAL VOC 2012 with 3D annotations in the form
of aligned 3D CAD models for 11 classes ($aeroplane$, $bicycle$,
$boat$, $bus$, $car$, $chair$, $diningtable$, $motorbike$, $sofa$,
$train$, and $tv$ $monitor$) of the $train$ and $val$ subsets.
The alignments are obtained through human supervision, by first selecting
the visually most similar CAD model for each instance, and specifying
the correspondences between a set of 3D CAD model keypoints and their
image projections, which are used to compute the 3D pose of the instance
in the image. The rich object annotations include object pose and
shape, and we use them as a test bed for our analysis. Unless otherwise
stated all presented models are trained on the Pascal3D+ $train$
set and evaluated on its test set (Pascal VOC 2012 $val$).

\section{\label{sec:Synthetic-images}Synthetic images}

\Cnn s reach high classification/detection quality by using a large
parametric model (e.g. in the order of $10^{7}$ parameters). The
price to pay is that \cnn s need a large training set to reach top
performance. We want to explore whether the performance scales as
we increase the amount of training data. To that end, we explore two
possible directions to increase the data volume: data augmentation
and synthetic data generation. 

Data augmentation consists of creating new training samples by simple
transformations of the original ones (such as scaling, cropping, blurring,
subtle colour shifts, etc.), and it is a common practice during training
on large convnets \cite{Krizhevsky2012Nips,Chatfield2014Bmvc}. To
generate synthetic images we rely on CAD models of the object classes
of interest. Rendering synthetic data has the advantage that we can
generate large amounts of training data in a controlled setup, allowing
for arbitrary appearance factor distributions. For our synthetic data
experiments we use an extended set of CAD models, and consider multiple
types of renderings (\S\ref{sub:Rendering-types}).

\paragraph{\label{sec:cad-models}Extended Pascal3D+ CAD models }

Although the Pascal3D+ dataset \cite{xiang14wacv} comes with its
own set of CAD models, this set is rather small and it comes without
material information (only polygonal mesh). Thus the Pascal3D+ models
alone are not sufficient for our analysis. We extend this set with
models collected from internet resources. We use an initial set of
$\sim\negmedspace40$ models per class. For each Pascal3D+ training
sample we generate one synthetic version per model using a ``plain
texture'' rendering (see next section) with the same camera-to-object
pose. We select suitable CAD models by evaluating the R-CNN (trained
on Pascal 2007 train set) on the rendered images, and we keep a model
if it generates the highest scoring response (across CAD models) for
at least one training sample. This procedure makes sure we only use
CAD models that generate somewhat realistic images close to the original
training data distribution, and makes it easy to prune unsuitable
models. Out of $\sim\negmedspace440$ initial models, $\sim\negmedspace275$
models pass the selection process ($\sim\negmedspace25$ models per
class).
\begin{figure}[t]
\begin{centering}
\vspace{-1em}
\hspace*{\fill}\subfloat[\label{fig:real-data}Real image]{\protect\begin{centering}
\begin{tabular}{c}
\protect\includegraphics[width=0.23\columnwidth]{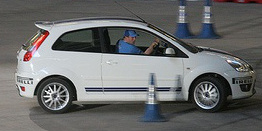}\tabularnewline
\protect\includegraphics[width=0.23\columnwidth]{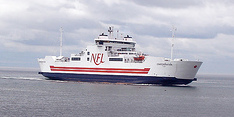}\tabularnewline
\end{tabular}\protect
\par\end{centering}

\protect\centering{}\protect}\hspace*{\fill}\subfloat[\label{fig:Wireframe}Wire-frame]{\protect\begin{centering}
\begin{tabular}{c}
\protect\includegraphics[width=0.23\columnwidth]{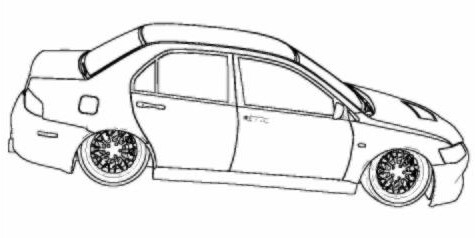}\tabularnewline
\protect\includegraphics[width=0.23\columnwidth]{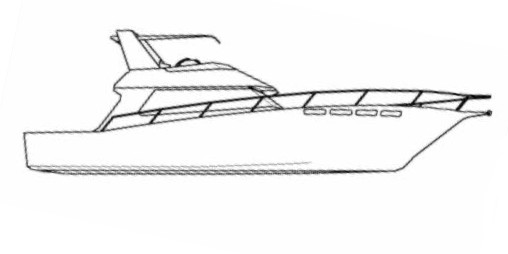}\tabularnewline
\end{tabular}\protect
\par\end{centering}

}\hspace*{\fill}\subfloat[\label{fig:plain-material}Plain texture]{\protect\begin{centering}
\begin{tabular}{c}
\protect\includegraphics[width=0.23\columnwidth]{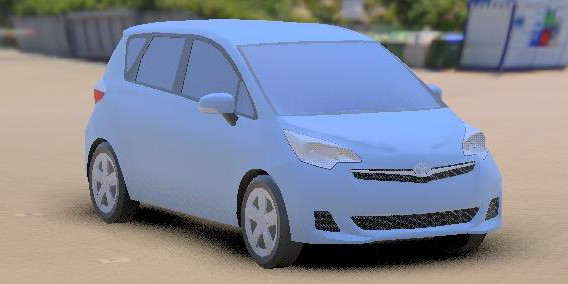}\tabularnewline
\protect\includegraphics[width=0.23\columnwidth]{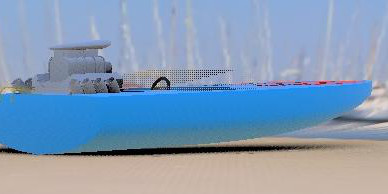}\tabularnewline
\end{tabular}\protect
\par\end{centering}

}\hspace*{\fill}\subfloat[\label{fig:texture-transfer}Texture transfer]{\protect\begin{centering}
\begin{tabular}{c}
\protect\includegraphics[width=0.23\columnwidth]{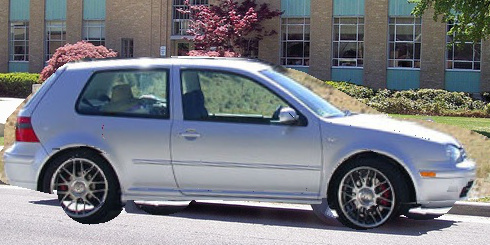}\tabularnewline
\protect\includegraphics[width=0.23\columnwidth]{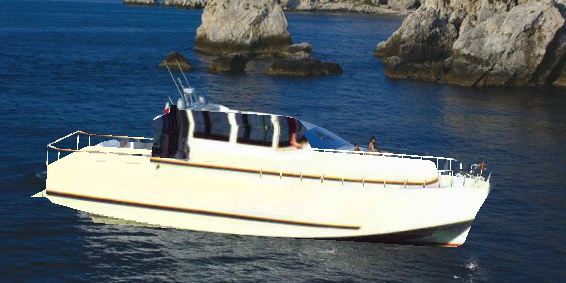}\tabularnewline
\end{tabular}\protect
\par\end{centering}

}\hspace*{\fill}
\par\end{centering}

\begin{centering}
\vspace{-0.5em}

\par\end{centering}

\protect\caption{\label{fig:rendering-types}Example training samples for different
type of synthetic rendering. Pascal3D+ training set.}
\vspace{-1em}
\end{figure}

\subsection{\label{sub:Rendering-types}Rendering types}

A priori it is unclear which type of rendering will be most effective
to build or augment a \cnn\xspace training set. We consider multiple
options using the same set of CAD models. Note that all rendering
strategies exploit the Pascal3D+ data to generate training samples
with a distribution similar to the real data (similar size and orientation
of the objects). See Fig. \ref{fig:rendering-types} for example renderings.

\paragraph{Wire-frame}

Using a white background, shape boundaries of a CAD model are rendered
as black lines. This rendering reflects the shape (not the mesh) of
the object, abstracting its texture or material properties and might
help the detector to focus on the shape aspects of the object.

\paragraph{Plain texture}

A somewhat more photo-realistic rendering considers the material
properties (but not the textures), so that shadows are present. We
considered using a blank background, or an environment model to generate
plausible backgrounds. We obtain slightly improved results using the
plausible backgrounds, and thus only report these results. This rendering
provides ``toy car'' type images, that can be considered as middle
ground between ``wire frame'' and ``texture transfer'' rendering.

\paragraph{Texture transfer}

All datasets suffer from bias \cite{Torralba2011CvprDatasetBias},
and it is hard to identify it by hand. Ideally, synthetic renderings
should have the same bias as the real data, while injecting additional
diversity. We aim at solving this by generating new training samples
via texture transfer. For a given annotated object on the Pascal3D+
dataset, we have both the image it belongs to and an aligned 3D CAD
model. We create a new training image by replacing the object with
a new 3D CAD model, and by applying over it a texture coming from
a different image. This approach allows to generate objects with slightly
different shapes, and with different textures, while still adequately
positioned in a realistic background context (for now, our texture
transfer approach ignores occlusions). This type of rendering is close
to photo-realistic, using real background context, while increasing
the diversity by injecting new object shapes and textures.

As we will see in \S \ref{sec:Which-synthetic-data-helps}, it turns
out that any of our renderings can be used to improve detection performance.
However the degree of realism affects how much improvement is obtained.
\vspace{-1em}

\section{\label{sec:What-did-the-convnet-learn}What did the network learn
from real data?\vspace{-0.5em}
}

In this section we analyze R-CNNs detection performance  in an attempt
to understand what have the models actually learned. We first explore
models performance across different appearance factors ($\S$\ref{sub:ap-vs-factor}),
going beyond the usual per-class detection performance. Second, we
dive deeper and aim at understanding what have the network layers
actually learned ($\S$\ref{sub:disentanglement}). \vspace{-1em}

\subsection{\label{sub:ap-vs-factor}Detection performance across appearance
factors}

To analyze the performance across appearance factors we split each
factor into equi-spaced bins. We present a new evaluation protocol
where for each bin only the data falling in it are actually considered
in the evaluation and the rest are ignored. This allows to dissect
the detection performance across different aspects of an appearance
factor. The original R-CNN\cite{Girshick2014ArxivRCNN} work includes
a similar analysis based on the toolkit from \cite{Hoiem2012EccvDiagnosing}.
Pascal3D+ however enables a more fine-grained analysis. Our experiments
report results for AlexNet (51.2 mAP)\cite{Krizhevsky2012Nips}, GoogleNet
(56.6 mAP)\cite{Szegedy2014ArxivGoogleNet}, VGG16 (58.8 mAP)\cite{simonyan15iclr}
and their combination (62.4 mAP).

\paragraph{Appearance factors}

We focus the evaluation on the following appearance factors: rotation
(azimuth, elevation), size, occlusion and truncation as these factors
have strong impact on objects appearance. Azimuth and elevation refer
to the angular camera position w.r.t. the object. Size refers to the
bounding box height. Although the Pascal3D+ dataset comes with binary
occlusion and truncation states, using the aligned CAD models and
segmentation masks we compute level of occlusion as well as level
and type of truncation. While occlusion and truncation levels are
expressed as object area percentage, we distinguish between 4 truncation
types: bottom (b), top (t), left (l) and right (r) truncation.

\paragraph{Analysis}

\begin{wrapfigure}{r}{0.33\columnwidth}%
\begin{centering}
\vspace{-0.5em}
\includegraphics[width=0.35\columnwidth]{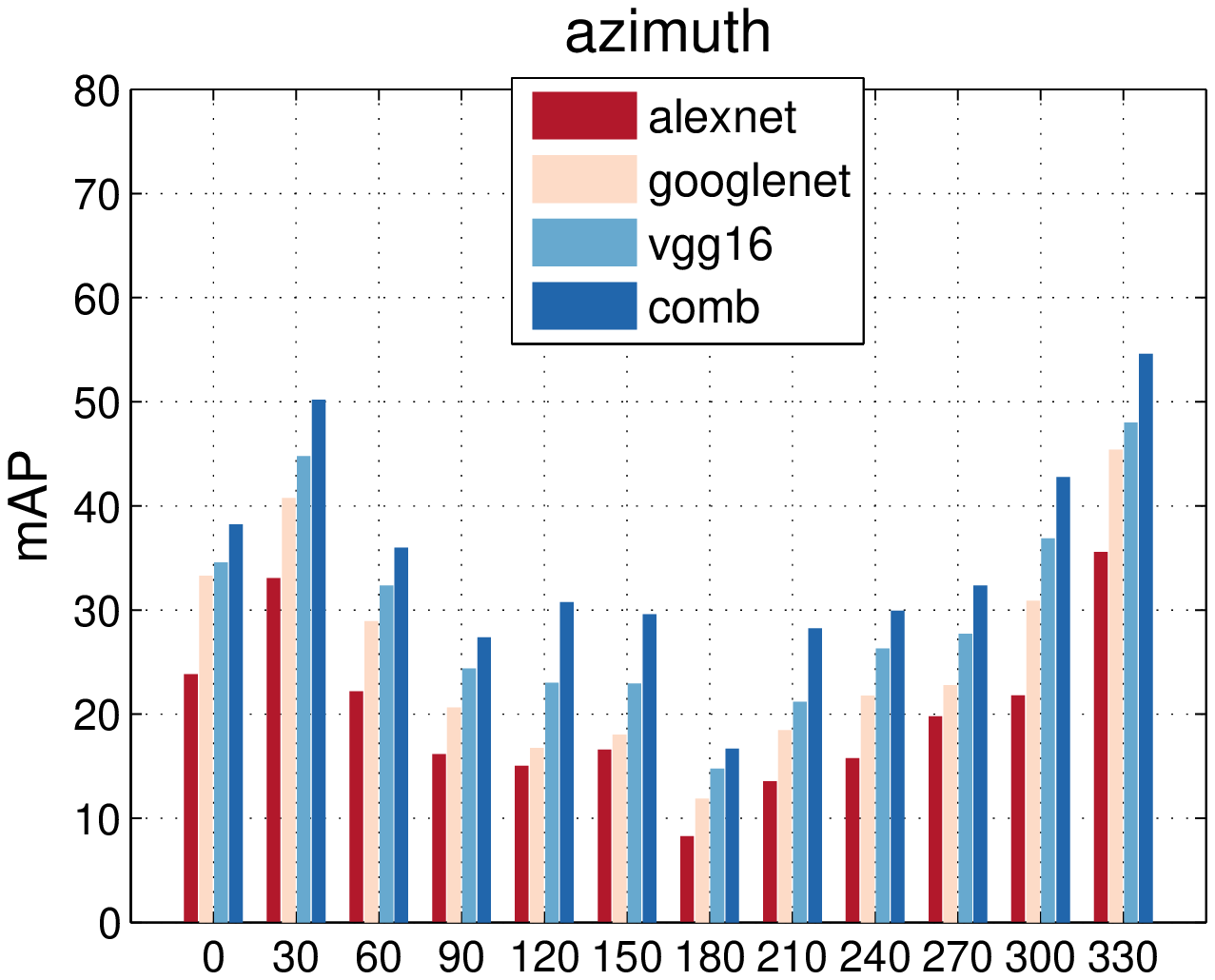}
\par\end{centering}

\begin{centering}
\includegraphics[width=0.35\columnwidth]{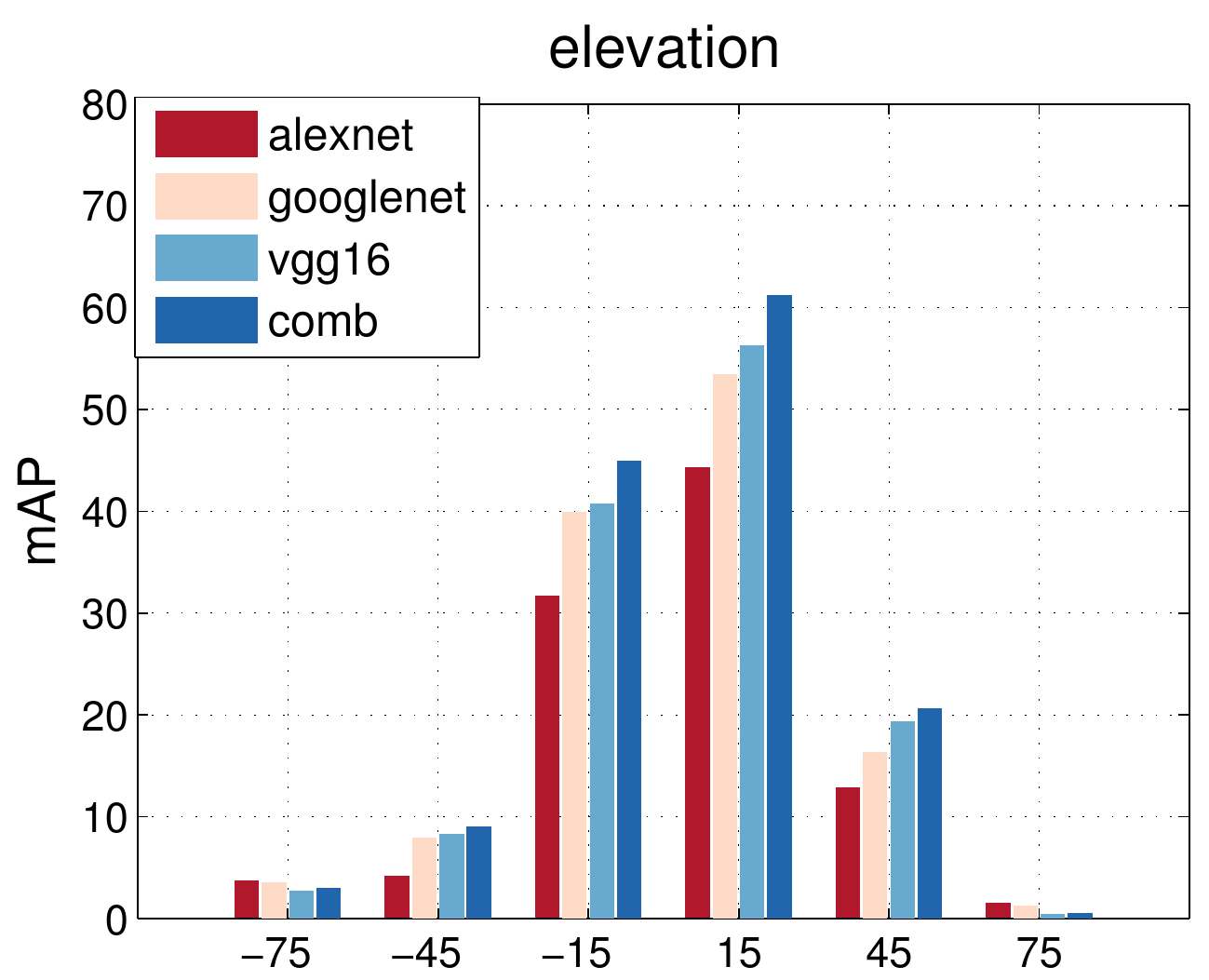}
\par\end{centering}

\begin{centering}
\includegraphics[width=0.35\columnwidth]{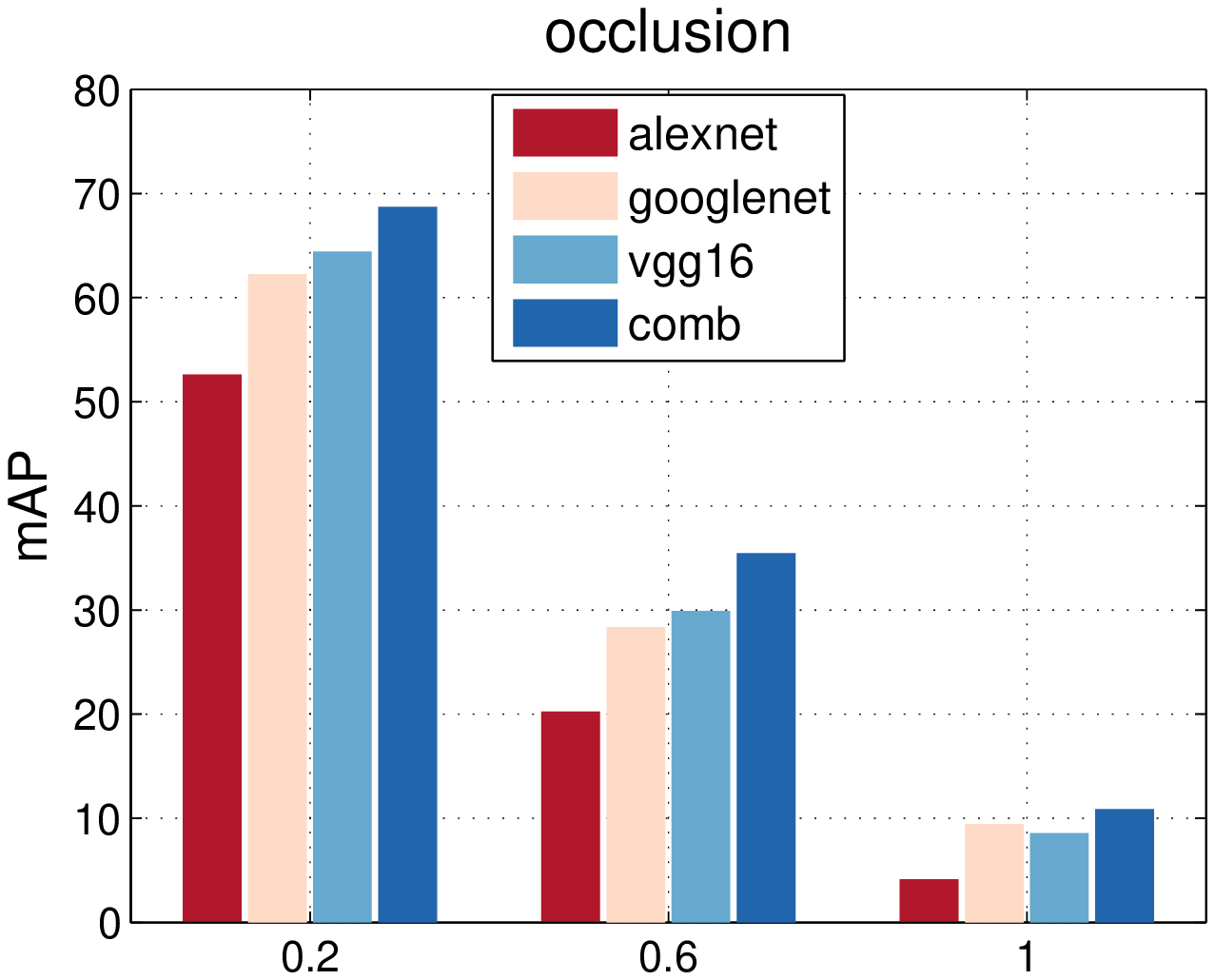}
\par\end{centering}

\begin{centering}
\includegraphics[width=0.35\columnwidth]{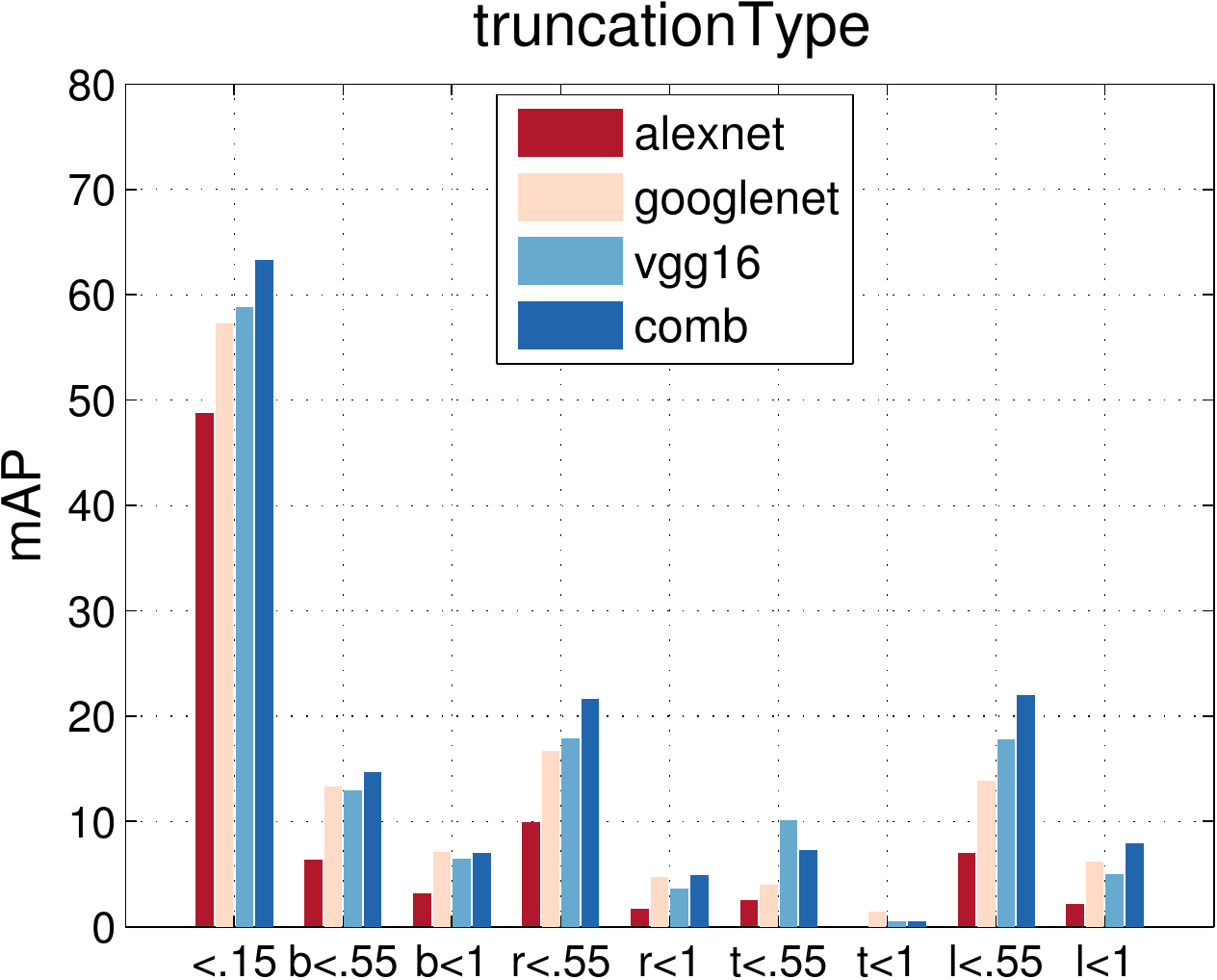}
\par\end{centering}

\begin{centering}
\includegraphics[width=0.35\columnwidth]{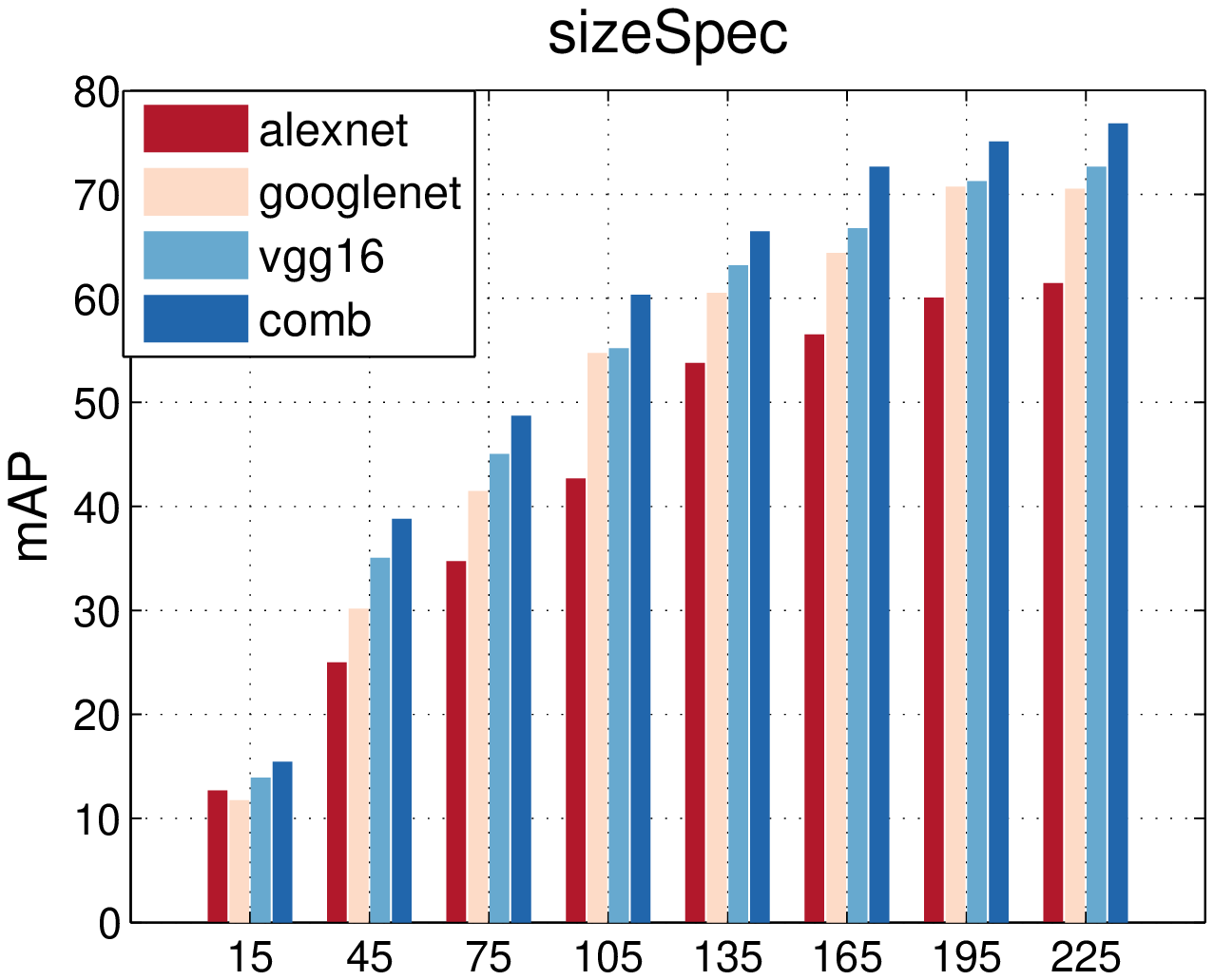}
\par\end{centering}

\protect\caption{\label{fig:ap-versus-factors}mAP of R-CNN over appearance factors.
 Pascal3D+ dataset.}
\vspace{-5em}
\end{wrapfigure}

Fig. \ref{fig:ap-versus-factors} reports performance across the
factors. The results point to multiple general observations. First,
there is a clear ordering among the models. VGG16 is better than GoogleNet
on all factor bins, which in turn consistently %
outperforms AlexNet. The combination of the three models (SVM trained
on concatenated features) consistently outperforms all of them suggesting
there is underlying complementarity among the networks. Second, the
relative strengths and weaknesses across the factors remain the same
across models. All networks struggle with occlusions, truncations,
and objects below 120 pixels in height. Third, for each factor the
performance is not homogeneous across bins, suggesting the networks
are not invariant w.r.t. the appearance factors.

It should be noted that there are a few confounding factors in the
results. First such factor is the image support (pixel area) of the
object, which is strongly correlated with performance. Whenever the
support is smaller e.g. small sizes, large occlusions/truncations
or frontal views the performance is lower. Second confounding factor
is the training data distribution. For a network with a finite number
of parameters, it needs to decide to which cases it will allocate
resources. The loss used during training will push the network to
handle well the most common cases, and disregard the rare cases. Typical
example is the elevation, where the models learn to handle well the
near $0^{\circ}$ cases (most represented), while they all fail on
the outliers: upper ($90^{\circ})$and lower ($-90^{\circ}$) cases.
We explore precisely this aspect in section \ref{sec:What-could-the-convnet-learn},
where we investigate performance under different training distributions.

\paragraph{Conclusion}

There is a clear performance ordering among the \cnn s which all
have similar weaknesses, tightly related to data distribution and
object area. Occlusion, truncation, and small size objects are clearly
weak points of the R-CNN detectors (arguably harder problems by themselves).
Given similar tendencies next sections focus on AlexNet.

\subsection{\label{sub:disentanglement}Appearance vector disentanglement}

\begin{figure}[t]
\begin{centering}
\vspace{-1em}
\hspace*{\fill}\subfloat[\label{fig:Class-disentanglement}Class]{\protect\begin{centering}
\protect\includegraphics[width=0.24\textwidth]{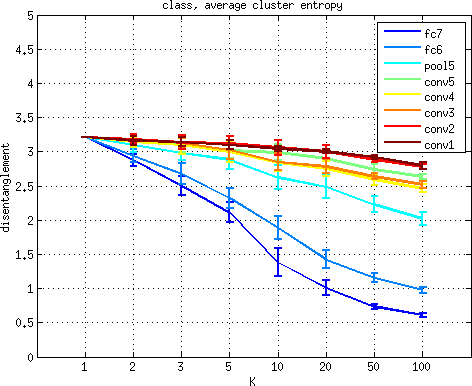}\protect
\par\end{centering}

}\hspace*{\fill}\subfloat[\label{fig:Azimuth-disentanglement}Azimuth]{\protect\begin{centering}
\protect\includegraphics[width=0.24\textwidth]{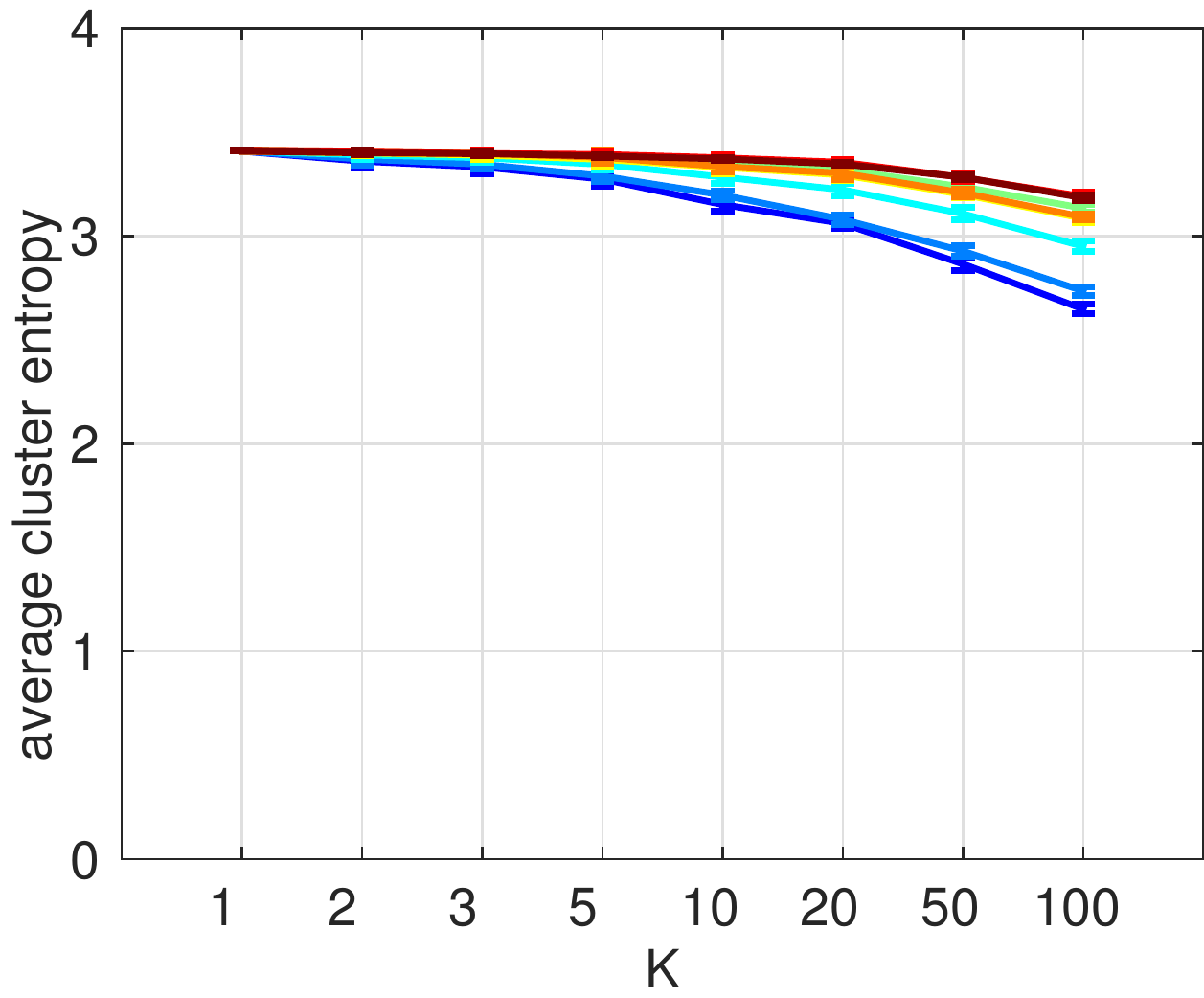}\protect
\par\end{centering}

}\hspace*{\fill}\subfloat[\label{fig:Elevation-disentanglement}Elevation]{\protect\begin{centering}
\protect\includegraphics[width=0.24\textwidth]{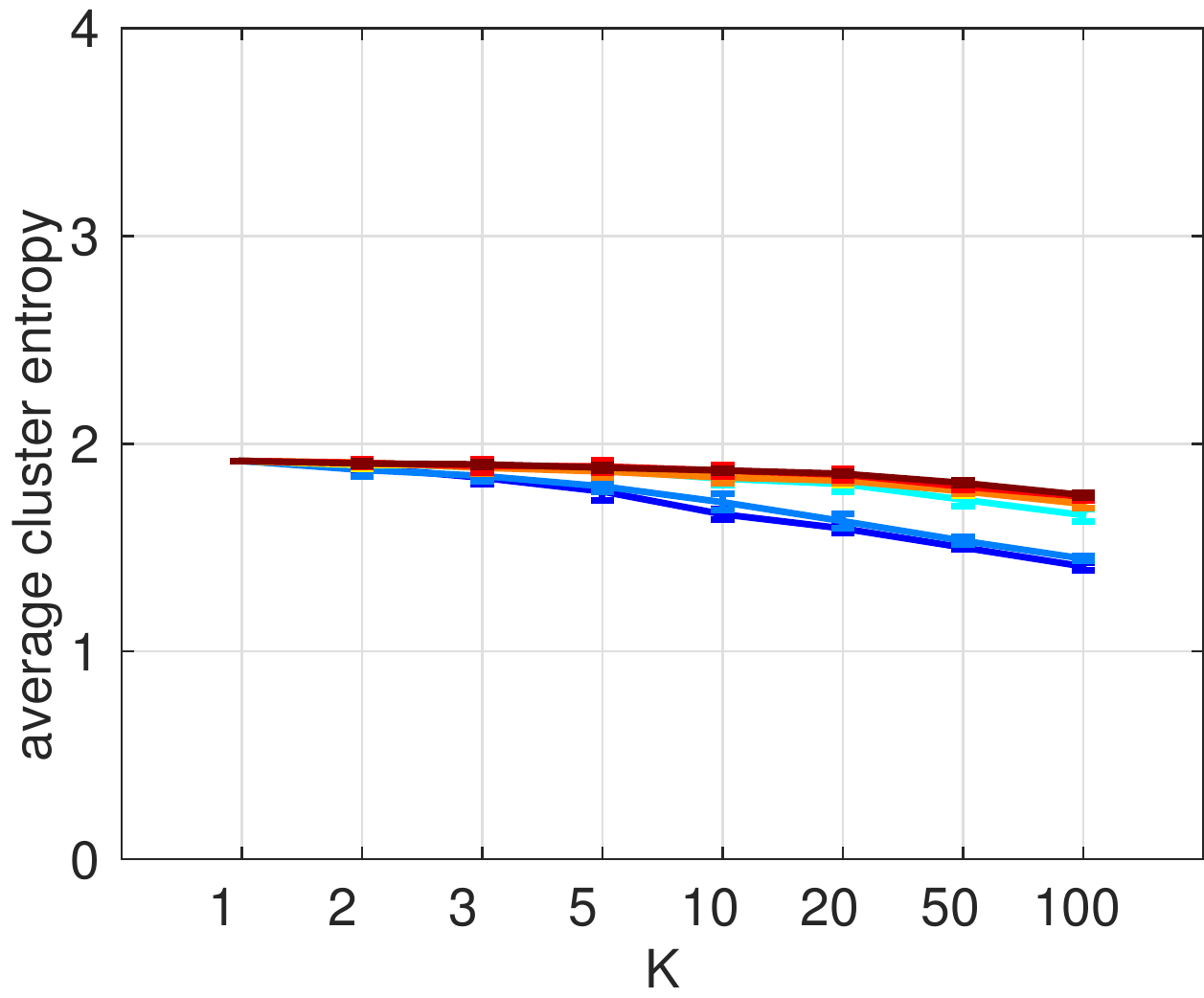}\protect
\par\end{centering}

}\hspace*{\fill}\subfloat[\label{fig:Shape-disentanglement}Shape]{\protect\centering{}\protect\includegraphics[width=0.24\textwidth]{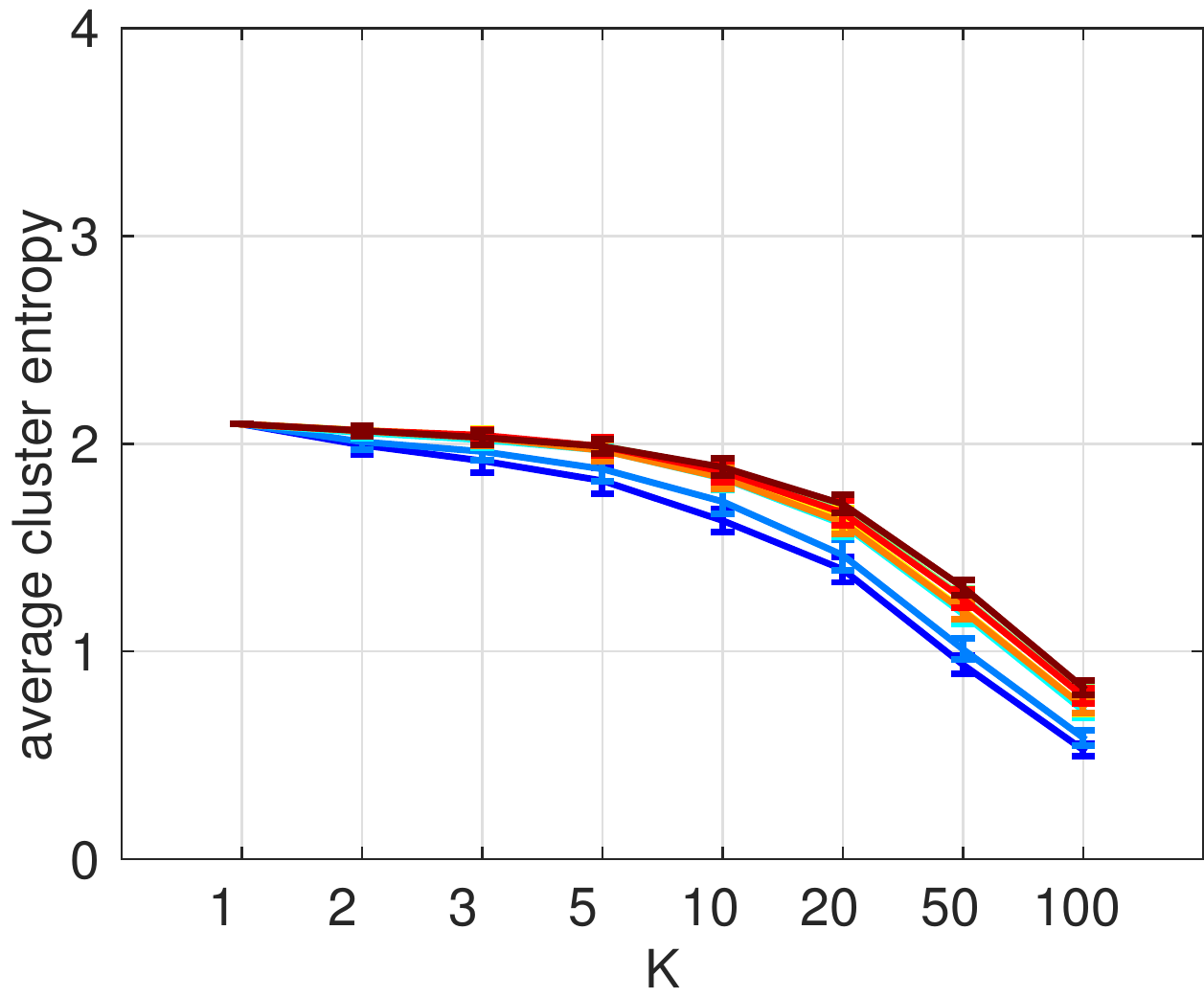}\protect}\hspace*{\fill}\vspace{-0.5em}

\par\end{centering}

\protect\caption{\label{fig:disentanglement}Average cluster entropy versus number
of clusters $K$; at different layers, for different appearance factors.
Pascal3D+ test data. }
\vspace{-1.5em}
\end{figure}

Other than just the raw detection quality, we are interested in understanding
what did the network learn internally. While previous work focused
on specific neuron activations \cite{goodfellow09nips}, we aim at
analyzing the feature representations of individual layers. Given
a trained network, we apply it over positive test samples, and cluster
the feature vectors at a given layer. We then inspect the cluster
entropy with respect to different appearance factors, as we increase
the number of clusters. The resulting curves are shown in Fig. \ref{fig:disentanglement}.
Lower average entropy indicates that at the given layer the network
is able to disentangle the considered appearance factor. Disentanglement
relates to discriminative power, invariance, and equivariance. (Related
entropy based metric is reported in \cite{Agrawal2014Eccv}, however
they focus on individual neurons).

\paragraph{Analysis}

From Fig. \ref{fig:Class-disentanglement} we see that classes are
well disentangled. As we go from the lowest conv1 layer to the highest
fc7 layer the disentanglement increases, showing that with depth the
network layers become more variant w.r.t. category. This is not surprising
as the network has been trained to distinguish classes. On the other
hand for azimuth, elevation and shape (class-specific disentanglement)
the disentanglement across layers and across cluster number stays
relatively constant, pointing out that the layers are not as variant
to these factors. We also applied this evaluation over plain texture
renderings (see \S\ref{sec:Synthetic-images} and \S\ref{sec:Which-synthetic-data-helps})
to evaluate the disentanglement of CAD models, the result is quite
similar to Fig. \ref{fig:Class-disentanglement}. 

\paragraph{Conclusion}

We make two observations. First, convnet representations at higher
layers disentangle object categories well, explaining its strong recognition
performance. Second, network layers are to some extent invariant to
different factors. \vspace{-2em}

\section{\label{sec:What-could-the-convnet-learn}What could the network learn
with more data?\vspace{-0.5em}
}

Section \ref{sec:What-did-the-convnet-learn} inspected what the network
learned when trained with the original training set. In this section
we explore what the network could learn if additional data is available.
We will focus on size (\S\ref{sub:Size-handling}), truncation (\S\ref{sub:Truncation-handling}),
and occlusion (\S\ref{sub:Occlusion-handling}) cases since these
are aspects that R-CNNs struggle to handle. For each case we consider
two general approaches: changing the training data distribution, or
using additional supervision during training. For the former we use
data augmentation to generate additional samples for specific size,
occlusion, or truncation bins. Augmenting the training data distribution
helps us realize if adding extra training data for a specific factor
bin helps improving the performance on that particular bin. When using
additional supervision, we leverage the annotations to train a separate
model for each bin. Providing an explicit signal during training forces
the network to distinguish among specific factor bins. The experiments
involve fine-tuning the R-CNN only (no SVM on top) as we are interested
in \cnn\xspace modelling capabilities.\vspace{-1em}
\begin{figure}[t]
\begin{centering}
\vspace{-1em}
\includegraphics[width=1\textwidth]{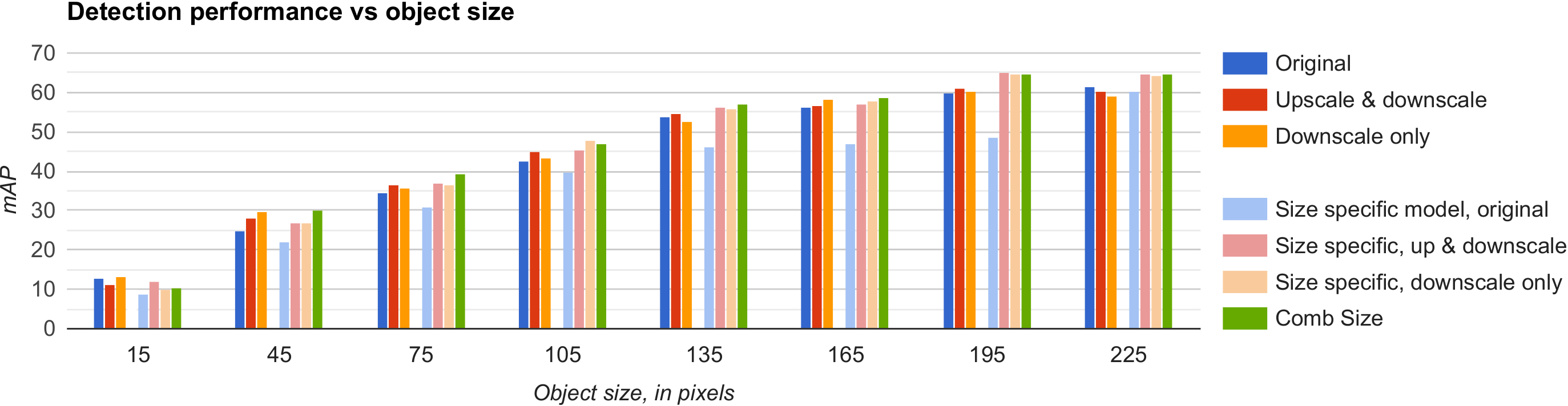}\vspace{-1em}

\par\end{centering}

\protect\caption{\label{fig:Training-for-size}Training with varying object size distribution.}
\vspace{-1.8em}
\end{figure}

\subsection{\label{sub:Size-handling}Size handling}

\begin{wrapfigure}{r}{0.5\columnwidth}%
\begin{centering}
\vspace{-4em}
\begin{minipage}[t]{0.5\textwidth}%
\begin{center}
\includegraphics[width=1\textwidth]{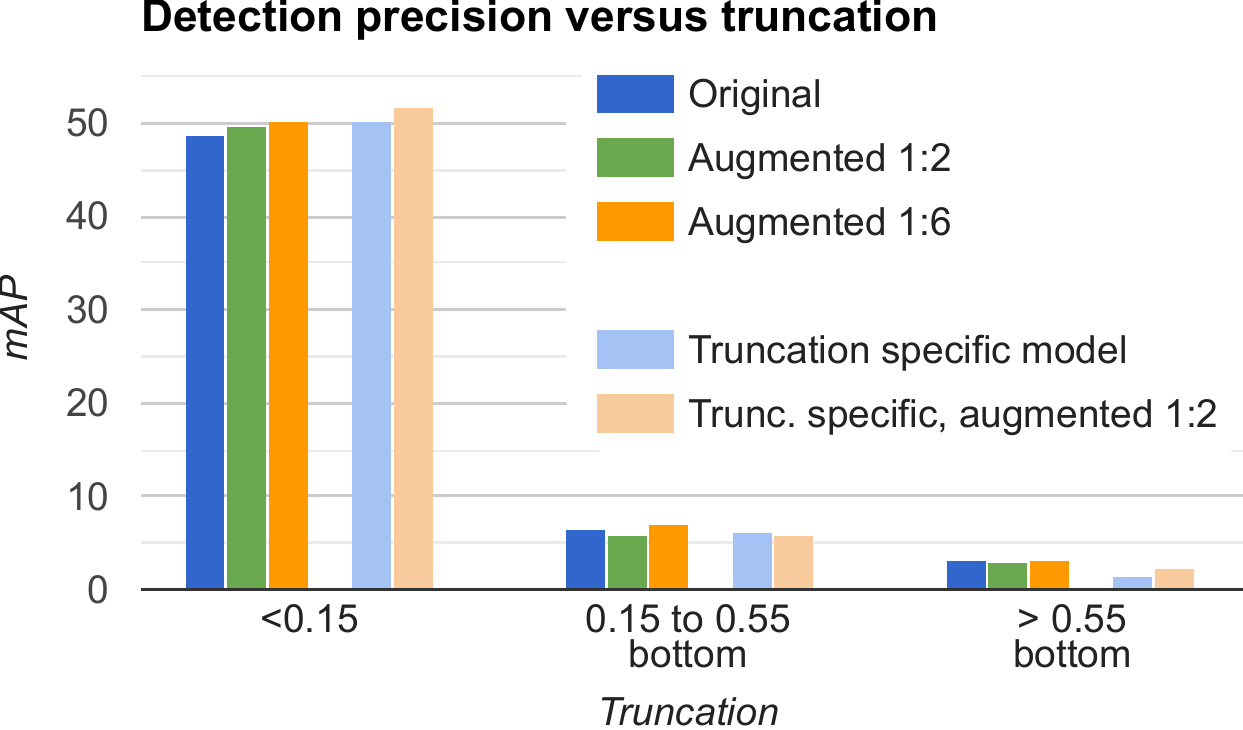}\vspace{-1.5em}

\par\end{center}

\protect\caption{\label{fig:Training-for-truncation}Training with varying truncated
objects distribution.}
\end{minipage}\vspace{-0em}

\par\end{centering}

\centering{}%
\begin{minipage}[t]{0.45\textwidth}%
\begin{center}
\includegraphics[width=0.9\textwidth]{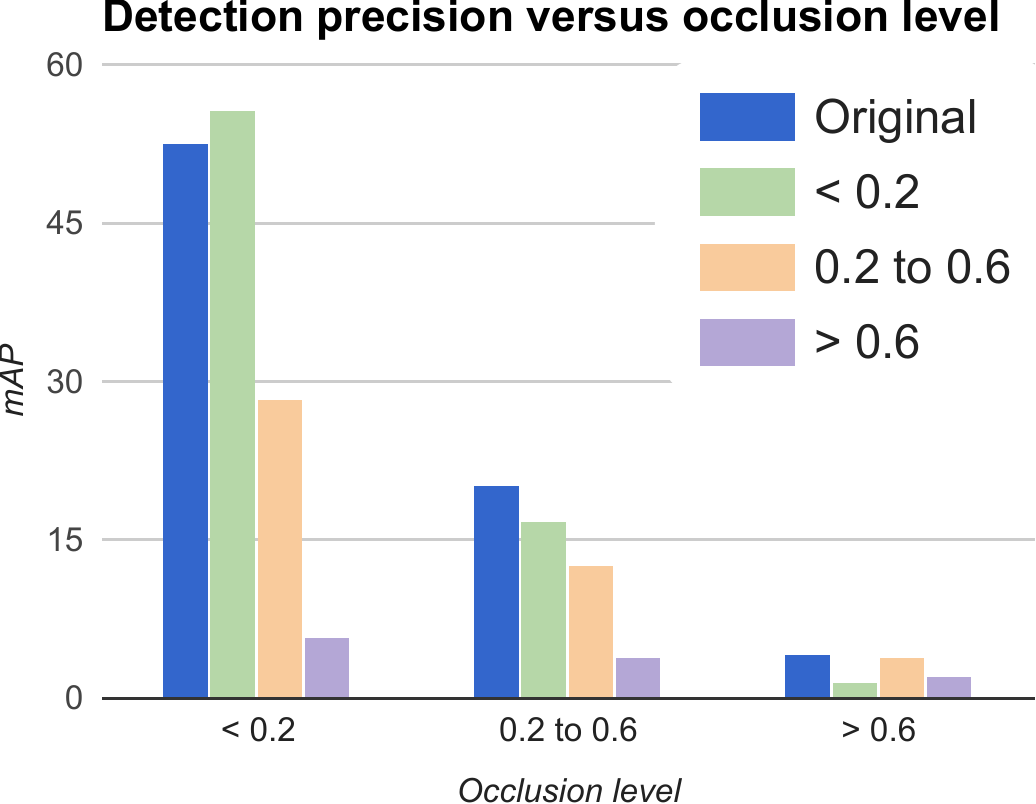}\vspace{-1.5em}

\par\end{center}

\protect\caption{\label{fig:Training-for-occlusion}Training with varying occluded
objects distribution.}
\end{minipage}\vspace{-3em}
\end{wrapfigure}%
Fig. \ref{fig:Training-for-size} shows the results with different
object size training distributions.\vspace{-0.3em}

\paragraph{More data}

The ``original'' bars correspond to the results in Fig. \ref{fig:ap-versus-factors}.
``Up \& downscale'' corresponds to training with a uniform size
distribution across bins by up/down-scaling all training samples to
all bins. As upscaled images will be blurry, ``downscale only''
avoids such blur, resulting in a distribution with more small size
training samples than larger sizes. Results in Fig. \ref{fig:Training-for-size}
indicate that data augmentation across sizes can provide a couple
of mAP points gain for small objects, however the network still struggles
with small objects, thus it is not invariant w.r.t. size despite the
uniform training distribution.

\paragraph{Bin-specific models}

The right side bars of Fig. \ref{fig:Training-for-size} show results
for bin-specific networks. Each bar corresponds to a model trained
and tested on that size range. Both augmentation methods outperform
the original data distribution on all size bins (e.g. at 195 pixels,
``up \& downscale'' improves by 5.2 mAP). In ``comb size'' we
combine the ``up \& downscale'' size specific models via an SVM
trained on their concatenated features. This results in superior
overall performance (54.0 mAP) w.r.t. the original data (51.2 mAP
with SVM).

\paragraph{Conclusion}

These results indicate that a) adding data uniformly across sizes
provides mild gains for small objects and does not result in size
invariant models, suggesting that the models suffer from limited capacity
and b) training bin-specific models results in better per bin and
overall performance.\vspace{-1em}

\subsection{\label{sub:Truncation-handling}Truncation handling}

\paragraph{More data}

Fig. \ref{fig:Training-for-truncation} shows that generating truncated
samples from non-truncated ones, respecting the original data distribution,
help improve (1.5 mAP points) handling objects with minimal truncation;
but does not improve medium or large truncation handling (trend for
top, left and right is similar to the shown bottom case). Using biased
distributions provided worse results.

\paragraph{Bin-specific models}

Similar to the ``more data'' case, training a \cnn\xspace for each
specific truncation cases only helps for the low truncation cases,
but is ineffective for medium or large truncations.

\paragraph{Conclusion}

These results are a clear indication that training data do not help
per-se handling this case. Architectural changes to the detector seem
required to obtain a meaningful improvement.\vspace{-1em}

\subsection{\label{sub:Occlusion-handling}Occlusion handling}

Similar to the truncation case, Fig. \ref{fig:Training-for-occlusion}
shows that specialising a network for each occlusion case is only
effective for the low occlusion case. Medium/high occlusion cases
are a ``distraction'' for training non-occluded object detection.

\paragraph{Conclusion}

For truncation and occlusion, it seems that architectural changes
are needed to obtain significant improvements. Simply adding training
data or focusing the network on sub-tasks seems insufficient.

\section{\label{sec:Which-synthetic-data-helps}Does synthetic data help?}

\begin{wraptable}{r}{0.45\columnwidth}%
\begin{centering}
\vspace{-6.5em}
\begin{tabular}{ccc}
\textbf{Synthetic} & \textbf{Ratio } & \multirow{2}{*}{\textbf{mAP}}\tabularnewline
\textbf{type} & \hspace*{-0.25em}{\scriptsize{}Real:Synth.}\hspace*{-0.25em} & \tabularnewline
\midrule 
- & 1:0 & 47.6\tabularnewline
Wire-frame & 0:1 & 21.8\tabularnewline
Plain texture & 0:1 & 23.5\tabularnewline
Texture transfer & 0:1 & 38.4\tabularnewline
Wire-frame & 1:2 & 48.3\tabularnewline
Plain texture & 1:2 & 49.9\tabularnewline
Texture transfer & 1:2 & \textbf{51.5}\tabularnewline
\end{tabular}\vspace{-0.5em}

\par\end{centering}

\protect\caption{\label{tab:ap-vs-synth-type}Results with different synthetic data
type. Pascal3D+ test.}
\vspace{-2em}
\end{wraptable}%
We have seen that \cnn s have weak spots for object detection, and
adding data results in limited gains. As \cnn s are data hungry methods,
the question remains what happens when more data from the same training
distribution is introduced. Obtaining additional annotated training
data is expensive, thus we consider the option of using renderings.
Our results with renderings (see $\S$\ref{sec:Synthetic-images})
are summarised in Tab.~\ref{tab:ap-vs-synth-type}. Again we focus
on fine-tuning \cnn s only. All renderings are done using a similar
data distribution as the original one, aiming to improve on common
cases.

\paragraph{Analysis}

From Tab.~\ref{tab:ap-vs-synth-type} we observe that using synthetic
data alone (0:1 ratio) under-performs compared to using real data,
showing there is still room for improvement on the synthetic data
itself. That being said, we observe that even the arguably weak wire-frame
renderings do help improve detections when used as an extension of
the real data. We empirically chose data ratio of 1:2 between real
and synthetic as that seemed to strike good balance among the two
data sources. As expected, the detection improvement is directly
proportional to the photo-realism (see Tab.~\ref{tab:ap-vs-synth-type}).
This indicates that further gains can be expected as photo-realism
is improved. Our texture transfer approach is reasonably effective,
with a 4 mAP points improvement. Wire-frame renderings inject information
from the extended CAD models. The plain texture renderings additionally
inject information from the material properties and the background
images. The texture transfer renderings use Pascal3D+ data, which
include some ImageNet images too. If we add these images directly
to the training set (instead of doing the texture transfer) we obtain
50.6 mAP (original to additional ImageNet images ratio is 1:3). This
shows that the increased diversity of our synthetic samples further
help improving results. Plain textures provide 2 mAP points improvement,
and texture transfer 4 mAP points. In comparison, \cite{Girshick2015ArxivFastRcnn}
reports 3 mAP points gain (on Pascal VOC 2012 test set) when using
the Pascal VOC 2007 together with the 2012 training data (over an
R-CNN variant). Our gains are quite comparable to such number, despite
relying on synthetic renderings.

\paragraph{Conclusion}

\begin{wraptable}{r}{0.45\columnwidth}%
\begin{centering}
\vspace{-2em}
\begin{tabular}{cccc}
\textbf{Data} & \textbf{CNN} & \textbf{mAP} & \textbf{AAVP}\tabularnewline
\midrule
\multirow{4}{*}{{\scriptsize{}Pascal3D+}} & AlexNet & 51.2 & 35.3\cite{bojan15w3dsi}\tabularnewline
 & GoogleNet & 56.6 & -\tabularnewline
 & VGG16 & 58.8 & -\tabularnewline
 & comb & 62.6 & -\tabularnewline
\midrule
\multirow{2}{*}{{\scriptsize{}Pascal3D+}} & AlexNet & 54.6 & -\tabularnewline
 & GoogleNet & 59.1 & -\tabularnewline
{\scriptsize{}\&} & VGG16 & 61.9 & -\tabularnewline
{\scriptsize{}Texture} & comb & 64.1 & \textbf{43.8}\tabularnewline
{\scriptsize{}transfer} & comb+size & 64.7 & -\tabularnewline
 & {\small{}comb+bb} & 66.3 & -\tabularnewline
 & {\footnotesize{}\hspace*{-1em}comb+size+bb} & \textbf{67.2} & -\tabularnewline
\end{tabular}\vspace{-0.5em}

\par\end{centering}

\protect\caption{\label{tab:combine-networks}Pascal3D+ results}
\vspace{-4em}
\end{wraptable}%
Synthetic renderings are an effective mean to increase the overall
detection quality. Even simple wire-frame renderings can be of help.\vspace{-1em}

\section{All-in-one}

In Tab.~\ref{tab:combine-networks} we show results when training
the SVM on top of the concatenated features of the \cnn s fine-tuned
with real and mixed data. We also report joint object localization
and viewpoint estimation results (AAVP \cite{bojan15w3dsi} measure).
As in \cite{bojan15w3dsi}, for viewpoint prediction we rely on a
regressor trained on \cnn\xspace features fine-tuned for detection. 

We observe that the texture renderings improve performance on all
models (e.g. VGG16 58.8 to 61.9 mAP). Combining these three models
further improves the detection performance and achieves state-of-the-art
viewpoint estimation. Adding size specific VGG16 models (like in
\S\ref{sub:Size-handling}) further pushes the results, improving
(up to 5 mAP points) on small/medium sized objects. Adding bounding
box regression, our final combination achieves 67.2 mAP, the best
reported detection result on Pascal3D+. 

\vspace{-1em}

\section{\label{sec:Conclusion}Conclusion}

We presented new results regarding the performance and potential of
the R-CNN architecture. Although higher overall performance can be
reached with deeper \cnn s (VGG16), all the considered state-of-the-art
networks have similar weaknesses; they underperform for truncated,
occluded and small objects (\S\ref{sec:What-could-the-convnet-learn}).
Additional training data does not solve these weak points, hinting
that structural changes are needed. Despite common belief, our results
suggest these models are not invariant to various appearance factors.
Increased training data, however, does improve overall performance,
even when using synthetic image renderings (\S\ref{sec:Which-synthetic-data-helps}).

In future work, we would like to extend the CAD model set in order
to cover more categories. Understanding which architectural changes
will be most effective to handle truncation, occlusion, or small objects
remains an open question.

\newpage{}

\bibliographystyle{splncs03}
\bibliography{2015_iccv_rcnn_analysis}

\end{document}